\documentclass[letterpaper, 10 pt, conference]{ieeeconf}  % Comment this line out if you need a4paper

\IEEEoverridecommandlockouts                              % This command is only needed if you want to use the \thanks command

\usepackage{adjustbox}

\usepackage{duckuments}

\usepackage{multicol}
\usepackage{caption}
\usepackage{blindtext}
\usepackage{graphicx}
\usepackage{hyperref}
\usepackage{amsmath}
\usepackage{amssymb}
\usepackage{tabularx}
\usepackage{subcaption}
\usepackage[acronym]{glossaries}
\usepackage{wrapfig,lipsum,booktabs}
\usepackage{siunitx}
\usepackage{float}
\usepackage{cleveref}

\usepackage{multirow}
\usepackage[normalem]{ulem}
\useunder{\uline}{\ul}{}

\usepackage{array}
\usepackage{bbold}

\newacronym{dl}{DL}{Deep Learning}
\newacronym{gat}{GAT}{Graph Attention Network}
\newacronym{gnn}{GNN}{Graph Neural Network}
\newacronym{ccg}{CCG}{Corner Case Generation}
\newacronym{hgnn}{HGNN}{Heterogeneous Graph Neural Network}

\newacronym{gcn}{GCN}{Graph Convolutional Network}
\newacronym{rgcn}{RGCN}{Relational Data with Graph Evolutionary Networks}
\newacronym{hin}{HIN}{Heterogeneous Information Network}
\newacronym{auc}{AUC}{Area Under Curve}
\newacronym{roc}{ROC}{Receiver Operating Characteristic}
\newacronym{tpr}{TPR}{True Positive Rate}
\newacronym{fpr}{FPR}{False Positive Rate}
\newacronym{elu}{ELU}{Exponential Linear Unit}
\newacronym{npc}{NPC}{Non-Player Character}
\newacronym{sgg}{SGG}{Scene Graph Generation}
\newacronym{bp}{BP}{back-propagation}
\newacronym{hgn}{HGN}{Heterogeneous Graph Attention Network}
\newacronym{pyg}{PyG}{PyTorch Geometric}
\newacronym{mlp}{MLP}{Multi-Layer Perceptron}
\newacronym{adam}{Adam}{Adaptive Moment Estimation}
\newacronym{nhtsa}{NHTSA}{National Highway Traffic Safety Administration}
\newacronym{mlm}{MLM}{Masked Language Model}
\newacronym{gan}{GAN}{Generative Adversarial Net}
\newacronym{vae}{VAE}{Variational Autoencoders}
\newacronym{rl}{RL}{Reinforcement Learning}
\newacronym{leakyrelu}{LeakyReLU}{Leaky Rectified Linear Unit}
\newacronym{ad}{AD}{Autonomous Driving}
\newacronym{av}{AV}{Autonomous Vehicle}
\newacronym{llm}{LLM}{Large Language Model}
\newacronym{gru}{GRU}{Gated Recurrent Unit}
\newacronym{ragas}{RAGAs}{Retrieval Augmented Generation Assessment}
\newacronym{road}{ROAD}{ROad event Awareness Dataset}
\newacronym{bce}{BCE}{Binary Cross Entropy}
\newacronym{dtsg}{DTSG}{Dynamic Temporal Scene Graph}

\usepackage{amsmath}
\usepackage{amssymb}
\usepackage{mathtools}
\DeclarePairedDelimiterX\set[1]\lbrace\rbrace{\def\given{\;\delimsize\vert\;}#1}

\usepackage[hang,flushmargin]{footmisc}

\begin{document}

\title{\LARGE \bf
GraphSCENE: On-Demand Critical Scenario Generation \\ for Autonomous Vehicles in Simulation
}
\author{Efimia Panagiotaki$^{12}$, Georgi Pramatarov$^{1}$, Lars Kunze$^{23}$, Daniele De Martini$^{1}$ \\ 
$^{1}$Mobile Robotics Group (MRG) and $^{2}$Cognitive Robotics Group (CRG), University of Oxford, UK \\
$^{3}$Bristol Robotics Laboratory (BRL), University of the West of England, UK \\
{\texttt{\{efimia,georgi,lars,daniele\}@robots.ox.ac.uk}}
\thanks{The project was supported by the EPSRC Programme Grant ``From Sensing to Collaboration'' (EP/V000748/1), the EPSRC project RAILS (EP/W011344/1), and the Oxford Robotics Institute research project RobotCycle. Efimia Panagiotaki was supported by the DeepMind Engineering Science Scholarship. 
}
}
\maketitle

%%%%%%%%%%%%%%%%%%%%%%%%%%%%%%%%%%%%%%%%%%%%%%%%%%%%%%%%%%%%%%%%%%%%%%%%%%%%%%%%
\begin{abstract}
Testing and validating \gls{av} performance in safety-critical and diverse scenarios is crucial before real-world deployment. However, manually creating such scenarios in simulation remains a significant and time-consuming challenge. This work introduces a novel method that generates dynamic temporal scene graphs corresponding to diverse traffic scenarios, \textit{on-demand}, tailored to user-defined preferences, such as \gls{av} actions, sets of dynamic agents, and criticality levels. A temporal \gls{gnn} model learns to predict relationships between ego-vehicle, agents, and static structures, guided by real-world spatiotemporal interaction patterns and constrained by an ontology that restricts predictions to semantically valid links. Our model consistently outperforms the baselines in accurately generating links corresponding to the requested scenarios. We render the predicted scenarios in simulation to further demonstrate their effectiveness as testing environments for \gls{av} agents.

\end{abstract}

\glsresetall

%%%%%%%%%%%%%%%%%%%%%%%%%%%%%%%%%%%%%%%%%%%%%%%%%%%%%%%%%%%%%%%%%%%%%%%%%%%%%%%%

\section{Introduction}

\begin{figure}[t]
    \centering
    \includegraphics[trim=15 110 420 40,clip,width=0.83\columnwidth]{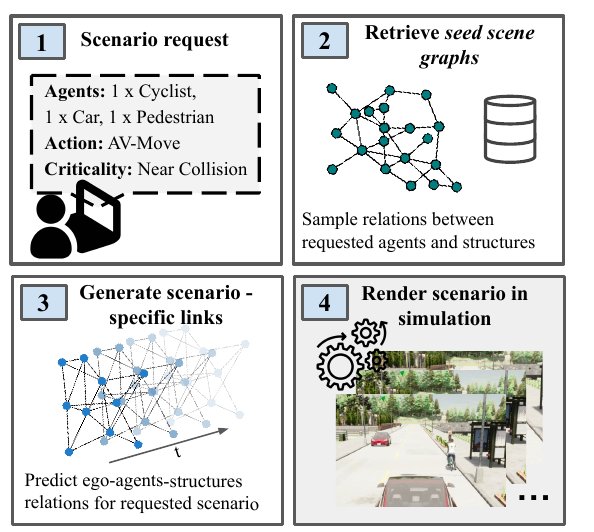}
\caption{\textit{On-demand} scenario generation pipeline: (1) User requests a scenario. (2) GraphSCENE samples relations between requested agents and static structures from a database, to create a scene graph used as \textit{seed} to (3) the temporal scene graph generation model, predicting scenario-specific links between the ego-vehicle node and all other graph entities. (4) The output scenario is rendered in simulation.}
    \label{fig:first_image}
    \vspace{-15pt}
\end{figure}

Ensuring the safety of passengers and road users through extensive testing and validation of \gls{ad} systems is an important task. Despite the availability of diverse datasets \cite{liu2021survey}, corner-case scenarios -- i.e. rare or unique situations \cite{breitenstein2020systematization} -- remain a significant challenge for \gls{av} systems \cite{sun2021corner}. These safety-critical scenarios often fall within the long tail of real-world dataset distributions \cite{zhou2022longtail}, as their unexpected occurrence makes them difficult to capture. Simulation has become an essential tool to identify potential failure points before real-world deployment through extensive \gls{av} testing in controllable, complex, diverse, and realistic traffic situations.

Effective scenario generation must capture complex, unsafe, and unpredictable interactions that \glspl{av} might encounter. Recent works range from leveraging real-world data to synthesising rare events in simulation, aiming to generate targeted scenarios for systematic safety validation, thus reducing the likelihood of unforeseen failures in \gls{av} deployment \cite{tan2023languageconditionedtrafficgeneration, feng2022trafficgen}. However, despite their promising results, data-driven approaches don't allow situation control, while simulation-based methods lack plausibility \cite{zhang2024chatscene}. Aiming to overcome these limitations, this work proposes a hybrid approach that combines data-driven insights with logical constraints to generate synthetic scenarios in simulation and create more comprehensive testing frameworks. 

\begin{figure*}[ht]

\begin{subfigure}{\textwidth}
\centering
  \includegraphics[trim=5 240 50 0,clip,width=\textwidth]{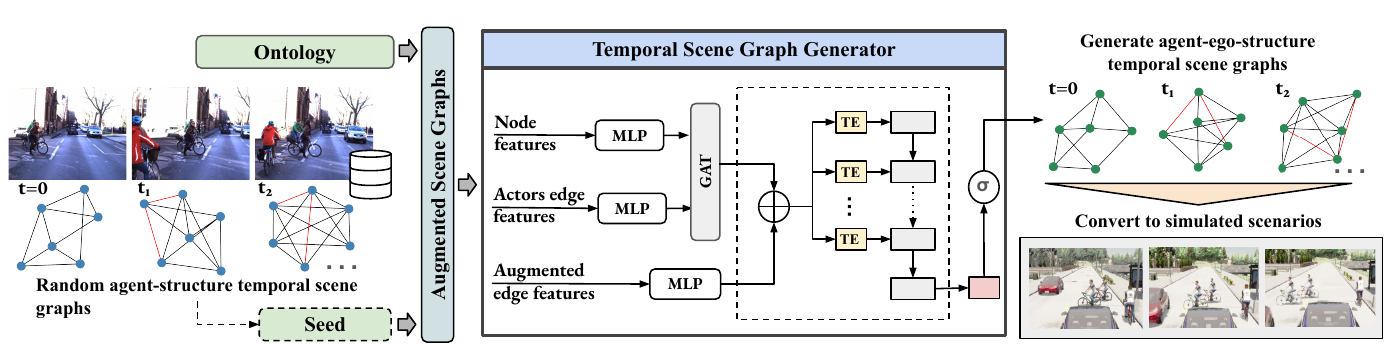}
  \vspace{-15pt}
\caption{}
\end{subfigure}

\begin{subfigure}{\textwidth}
\centering
  \includegraphics[trim=0 290 205 10,clip,width=0.75\textwidth]{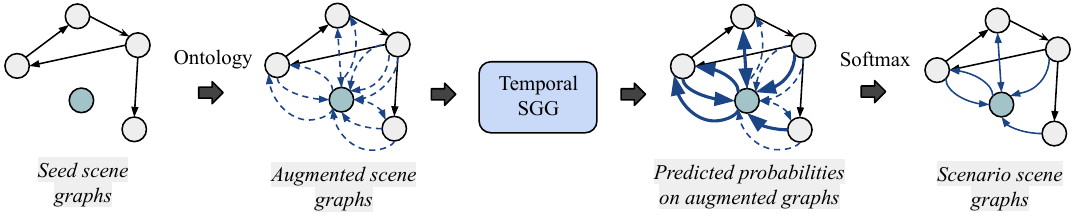}
  \vspace{-10pt}
  \caption{}
\end{subfigure}
   \caption{
Overview of our methodology. (a) We first extract temporal scene graphs from the \acrshort{road} dataset, capturing relations between agents and static structures, excluding the ego-vehicle. These serve as \textit{seeds} for a temporal scene graph generation model, which encodes heterogeneous features and, through a temporal encoder (TE), predicts relationships that semantically link the ego-vehicle with agents and static structures. The output, consisting of predicted \textit{scenario scene graphs}, models dynamic interactions and is converted into simulation scenarios for \gls{av} testing. (b) The underlying progression of link prediction, shown in a sample graph. Here, the blue nodes correspond to the ego-vehicle and grey nodes to the agents and static structures. We first augment the input \textit{seed scene graphs} with all possible relations from the ontology, allowing the temporal scene graph generation model to predict the likelihood of each edge in the augmented graphs. These edges correspond to edges in \textit{scenario scene graphs}.}

  \label{fig:overview}
  \vspace{-15pt}
\end{figure*}

Addressing the challenges of plausibility and diversity in synthetic datasets, our method generates safety-critical traffic scenarios \textit{on-demand} from user inputs, leveraging real-world data. The user simply requests a scenario and our model generates dynamic temporal scene graphs corresponding to the desired scenario, which are then rendered in CARLA Simulation \cite{carla}, as seen in \cref{fig:first_image}. In particular, the user defines the dynamic agents, the \gls{av} action (move, turn left/right, or overtake) and the scenario's criticality level (visible, near, or near collision).
Then, the relations between the agents -- excluding the \gls{av}, namely \textit{ego-vehicle} -- are sampled from our \textit{database} consisting of a collection of interactions extracted from the \gls{road} \cite{road}. These serve as \textit{seeds} for our graph generation model, which learns to generate temporal links corresponding to relations between the \gls{av} and the agents and static structures. This process is guided and constrained by an \textit{ontology} to ensure the plausibility of interactions and scene evolution. Our method provides a structured approach in generating complex traffic situations for evaluating \gls{av} navigation performance in diverse and complex scenarios.

Our key contributions are summarised as follows:
\begin{itemize}
    \item A novel end-to-end ontology-based and data-driven approach for \textit{on-demand} generation of safety-critical traffic scenarios in simulation;
    \item A novel learning methodology for temporal graph generation, capturing semantic, topological, and historical patterns between scene elements to generate realistic dynamic temporal scene graphs;
    \item A seamless integration of the generated scenarios into simulation, creating complex and diverse testing scenarios for evaluating \gls{av} navigation performance.
    
% The ability to seamlessly translate predicted scenarios into simulations, creating
%We show  that our predicted scenarios can be seamlessly translated into simulation in a real-to-sim fashion, creating realistic and plausible testing environments for \glspl{ad}.
\end{itemize}

% Quantitative and qualitative results affirm that our model successfully learns to generate driving scenarios both from real and random scenes, producing challenging scenarios in simulation for autonomous driving.

\section{Related Work}\label{sec:related_work}

Our method tackles \textit{traffic scenario generation} as a \textit{graph generation} task through \textit{temporal graph learning}.
This section reviews related work in these three key areas.

% Our method combines key concepts from multiple domains: real-to-sim synthetic data generation, traffic scenario generation, and temporal graph learning.

% \textbf{Real-to-Sim Synthetic Data Generation. }Several methods have been developed to create realistic synthetic datasets. Meta-Sim \cite{kar2019metasim} and Meta-Sim2 \cite{devaranjan2020metasim2} employ a meta-learning framework to optimise the simulator parameters using a differentiable simulation tuned on real-wold data.The real-to-sim generation network in \cite{prakash2021self} matches feature distributions between domains, facilitating both real-to-sim and sim-to-real transformations.
% Depth-SIMS \cite{musat2022depth} and NeuralFloors \cite{mucsat2024neuralfloors} generate high-fidelity images from real-world data but do not extend to full scenario generation. 
% In contrast, our approach generates dynamic, temporal scenarios in simulation based on real-world datasets.

\textbf{Traffic scenario generation. }Scenario generation is a key component of applications such as extracting scenarios from naturalistic driving data or adjusting scenario parameters \cite{schutt20231001}. However, existing methods are often limited in applicability and lack plausibility.
ChatScene \cite{zhang2024chatscene} employs a \gls{llm} to generate safety-critical scenarios in simulation, but is restricted by user prompts and lacks real-world grounding.
CC-SGG \cite{ccsgg} perturbs scene graphs to generate corner cases but relies on a small, simulation-based dataset and doesn't capture temporal patterns.
TrafficGen \cite{feng2022trafficgen} augments scenarios on HD maps and loads them in simulation. However, it relies on artificially generated agents, reducing scenario plausibility. Similarly, real-to-synthetic approaches \cite{tian2022real} represent scenes as graphs and generate actors' positions on HD maps to construct traffic scenarios, but do not directly render them in simulation. Rule-based methods \cite{kluck2018using,bogdoll2022ontology} generate scenarios via ontologies but often lack relevance to real traffic events. Our method overcomes these limitations by leveraging real-world data to capture complex interactions between scene participants and generate ontology-constrained temporal scene graphs, which are then rendered in simulation.

\textbf{Graph Generation. }Recent advancements in graph generation employ generative, autoregressive, or recurrent models to construct nodes and edges from incomplete input graphs. Variational Autoencoders (VAEs) have been employed to sequentially generate graph elements through a series of probabilistic decisions \cite{li2018learningdeepgenerativemodels}, but have applicability and scalability limitations. In protein structure prediction, \cite{NEURIPS2019_f3a4ff48} introduces an attention-based encoder with an autoregressive decoder, capturing node and edge features but not evolving relations over time. For scene graph generation, SceneGraphGen \cite{scenegraphgen2021} uses a \textit{seed} object in an autoregressive model but lacks temporal modeling, limiting its suitability for dynamic graphs. GraphRNN \cite{you2018graphrnngeneratingrealisticgraphs} treats graph generation as a sequential task, while GRAN \cite{liao2019gran} uses recurrent attention mechanisms to predict edges in augmented graphs. Despite strong performance, neither method captures evolving edge attributes over time. In comparison, our approach combines seed initialisation with edge prediction on augmented graphs, while uniquely capturing evolving spatiotemporal patterns. As such, our model captures temporal dependencies and intrinsic structural variations by generating links guided by the input seed graphs.

% Our approach builds on these foundations by focusing on scenario generation using dynamic temporal data from real-world datasets. We utilise the flexible structure of scene graphs to encode traffic agents and their interactions, enabling the prediction of interactions within traffic sequences. By conditioning on user-selected actions and criticalities, we generate realistic traffic scenarios in simulation, grounded in real-world data.

\textbf{Temporal Graph Learning. }Temporal graph learning has been applied to agent interactions modelling in traffic prediction.
STGCN \cite{stgnn} and ST-LGSL \cite{ST-LGSL} combine temporal and spatial aggregations for timeseries prediction, while ST-GAT \cite{stgat} captures spatiotemporal dependencies using graph attention and an LSTM.
Although effective, these methods mainly focus on specific traffic prediction tasks limiting their broader applicability.
Additionally, spatiotemporal scene graphs with LSTMs have been employed for explainable action prediction \cite{kochakarn2023explainable}, but they primarily capture temporal information at the node level rather than modeling the sequential evolution of events. Building on these foundations, our approach employs temporal \glspl{gnn} to encode traffic agents and model their evolving interactions, effectively generating plausible traffic scenarios.

%===============================================================================
\section{Overview}
\label{sec:overview}

Our approach, visualised in \cref{fig:overview}, uses scene graphs extracted from the \gls{road} dataset to generate new scenarios that satisfy user-defined requirements, such as configuration of agents, \gls{av} action, and scenario criticality. To achieve this, we first extract semantic elements from distinct frames in the ROAD dataset to build two types of graphs: (1) \textit{seed scene graphs} containing evolving relations between dynamic agents (excluding the ego-vehicle) and static structures, forming our \textit{database}, and (2) \textit{scenario scene graphs} including also relational information about the ego-vehicle.
% containing evolving relations between the ego-vehicle, dynamic agents, and static structures, corresponding to various scenarios.
The latter serve as ground truth and a weak supervision signal, as the model learns to generate links corresponding to scenario graphs, using the former as \textit{seeds}. Ultimately, our method predicts the relationships between the ego-vehicle and other scene entities to output dynamic scene graphs containing the desired temporal relations between scene elements, facilitating scenario generation in CARLA.

% To achieve this, our method predicts the relationships between the ego-vehicle and other scene entities that correspond to the desired scenario.

\subsection{Notation and Problem Definition}
We define a traffic scenario as $\mathbb{S} : \set{S_t \given t \in [1, N]}$ consisting of $N$ sequential scenes $S_t$.
Each $S_t$ is represented as a scene graph $G_t = \langle V_t, E_t \rangle$, where nodes $V_t$ represent key dynamic and static entities (e.g. cars, pedestrians, or lanes) and the edges $E_t$ denote their relationships (e.g. distance or topology).
Each node $n^i_t \in V_t$ corresponds to a semantic category (e.g., ego-vehicle, road, pedestrian, or cyclist) with a node feature representation $h_t^i$.
Similarly, each edge $e_t^{ij} \in E_t$ represents a semantic relationship (e.g., distance, topology) between two nodes $n^i_t, n_t^j \in V_t$ with an associate edge feature representation $p_t^{ij}$.
Edges can connect two entities or link an entity to itself, forming self-edges.
% A special node in each graph $G_t$ is the ego-vehicle, representing the \gls{av}.
% Such graphs are depicted in \cref{fig:scene_grahp}.
This creates a collection of $N$ graphs, $\mathbb{G} : \set{ G_t \given t \in [1, N] }$ for each scenario $\mathbb{S}$.
In our proposed method, the goal is to learn to generate \textit{scenario scene graphs}, $\mathbb{G}$, from \textit{seed scene graphs} $\tilde{\mathbb{G}}  : \set{ \tilde{G_t} \given t \in [1, N]}\text{, where each } \tilde{G_t} \subsetneq G_t$. The nodes $V_t$ and predicted relationships in $E_t$ from the generated scene graphs are used to create scenarios in simulation.

%===============================================================================
\section{Methodology}
\label{sec:methodology}
Our methodology consists of three main components: the extraction of the \textit{seed} and \textit{scenario scene graphs}, the temporal scene graph generation model, and the scenario integration in simulation. The following subsections detail our system.

\subsection{Scene Graph Extraction}

% shown in \cref{tab:ontology}, used to produce expressive and consistent graph representations of real traffic scenarios.

\textbf{Ontology Definition.} We establish a traffic domain \emph{ontology}, shown in \cref{tab:ontology}, structured around the Automotive Urban Traffic Ontology (A.U.T.O.) \cite{westhofen2022usingontologiesformalizationrecognition} and adapted for the \gls{road} dataset~\cite{road}. The ontology is used to retrieve nodes and edge labels for each graph $G_t$, integrating concepts such as object types, traffic rules, relative positions, actions, and behaviours in the graphs. The node labels are derived directly from the \gls{road} dataset~\cite{road} and consist of agents and locations -- (1) and (2) in \cref{tab:ontology}, respectively. Agent positions are represented by an \texttt{IsIn} relation linking to the locations in (2). Agent states are represented by self-relations, where for dynamic agents the states are in (3), for traffic lights the state is its colour $\{$\texttt{red, amber, green}$\}$, and for the ego-vehicle the state is in (4). Relative motion is captured by a relation in $\{$\texttt{MovingAway, MovingTowards}$\}$ to the ego node. 
We also include a \texttt{MustStop} traffic rule edge for agents, if the traffic light is red. 
Finally, we model the distance between agents and the ego-vehicle as a discrete proximity relation in (5).

\begin{table}[]
\vspace{7pt}
% \definecolor{lightgray}{rgb}{0.8, 0.8, 0.8}
\renewcommand{\arraystretch}{1.2}
\centering
% \begin{tabular}{l|l}
\begin{tabular}{p{1.8cm}|p{5.75cm}}
% \begin{tabular}{p{4cm}|p{12cm}}
& \textbf{Type} \\ \hline
(1) \textbf{Agents} & EGO, Pedestrian, Car, Cyclist, Motorbike, Bus, TrafficLight \\ \hline
(2) \textbf{Locations} & VehicleLane, OutgoingLane, OutgoingCycleLane, IncomingLane, IncomingCycleLane, Pavement, Junction, PedestrianCrossing, BusStop, Parking \\ \hline
(3) \textbf{Actions} & Move, Brake, Stop, IndicateLeft, IndicateRight, TurnLeft, TurnRight, Cross \\ \hline
(4) \textbf{AV Actions} & AV-Move, AV-MoveLeft, AV-MoveRight, AV-Overtake, AV-Stop, AV-TurnLeft, AV-TurnRight \\ \hline
(5) \textbf{Criticality} & Near Collision, Near, Visible \\ 
\end{tabular}
\caption{Entities (1)-(2) and relations (3)-(5) in our ontology forming \textit{node-edge-node} triplets.\label{tab:ontology}}
\vspace{-15pt}
\end{table}

\begin{figure}
\vspace{5pt}
\centering
\includegraphics[trim=0 500 0 0,clip,width=0.92\columnwidth]{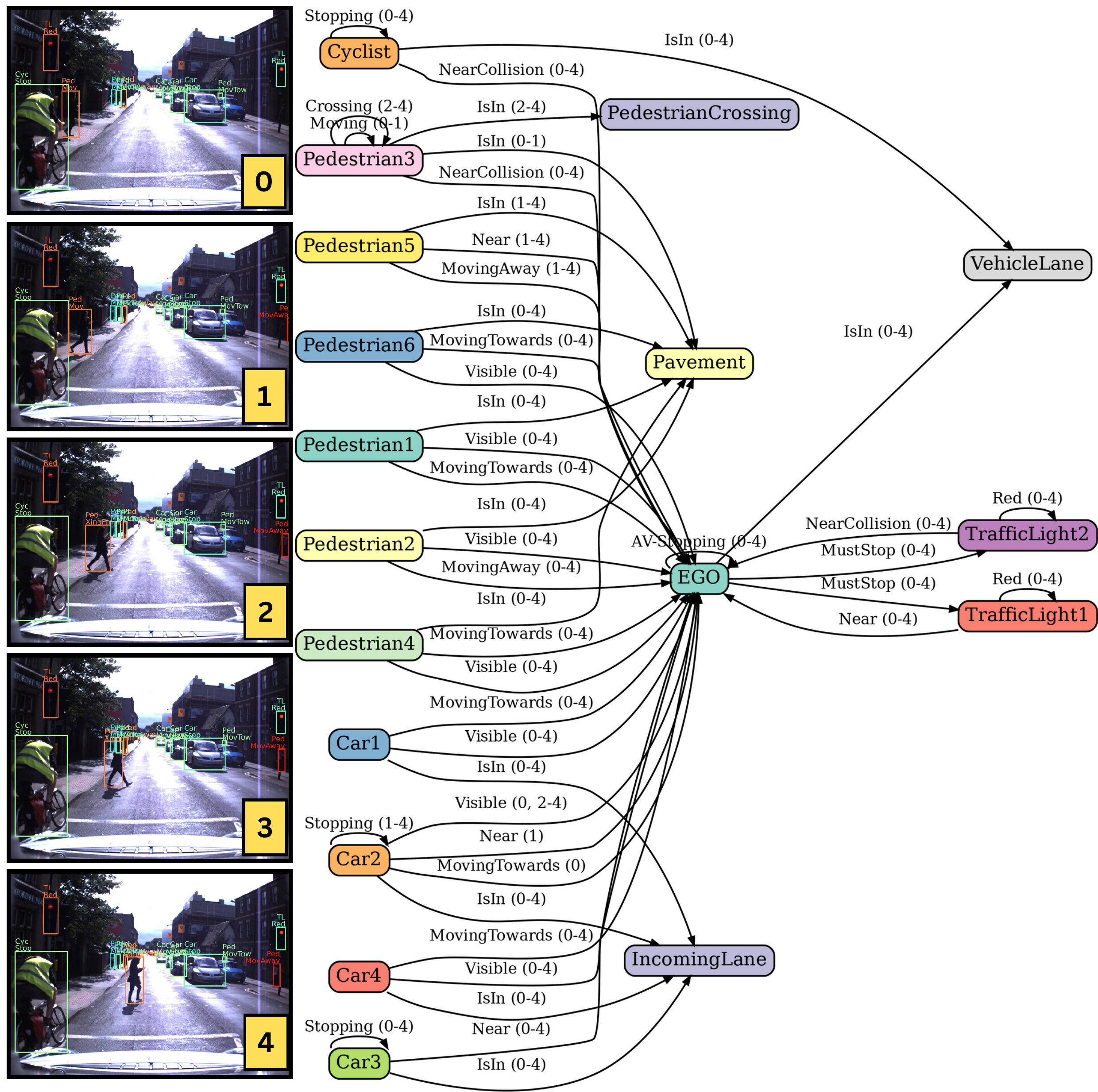}
 \caption{Example of scene graph extraction. A sequence of frames $\mathbb{S}$ is annotated (left) and a temporal scene graph $\mathbb{G}^T$ is constructed by linking the entities (right) using the ontology.}
\label{fig:scene_grahp}
\vspace{-15pt}
\end{figure}

    \textbf{Scenario Graphs.} For each frame in ROAD, we extract a graph $G_t$ where nodes represent entities and edges model the relations from the ontology. Their states and relationships are implemented as one-hot-encoded node and edge features. We extract a scenario $\mathbb{S}$ from a sequence of $N=5$ frames, where each frame corresponds to a scene $S_t$, that captures a single \gls{av} action, as seen in \cref{tab:ontology} (4). For each traffic scenario $\mathbb{S}$ and its corresponding scene graphs $\mathbb{G}$, we construct a temporal graph $\mathbb{G}^T = (\mathbb{V}^T, \mathbb{E}^T)$ by combining the nodes and edges from each $G_t \in \mathbb{G}$. The graphs $\mathbb{G}^T$ correspond to the temporal representation of the \textit{scenario scene graphs} $\mathbb{G}$.
Since objects -- i.e. agents and static structures -- can appear in multiple frames, we assign unique IDs and track them in order to only include unique nodes from $\mathbb{G}$ in $\mathbb{G}^T$. As multiple edges can be connected to the same node in $\mathbb{G}^T$, each edge is assigned a temporal label $\tau \in [0,4]$ depending on the graph $G_t$ it originated from, within the sequence $\mathbb{G}$. This process is illustrated in \cref{fig:scene_grahp}.

We then label the graph $\mathbb{G}^T$ with the ego-vehicle \gls{av} action and determine the scenario's \textit{criticality} level based on the proximity relations of all agents. Specifically, we assign the criticality level by selecting the most severe proximity relation label present in any graph $G_t \in \mathbb{G}$. The ego-vehicle \gls{av} action is derived from the last recorded \gls{av} action in $\mathbb{G}^T$. For example, if the ego-vehicle moves forward and then turns left, the scenario is labeled as `AV-TurnLeft'.
% Moreover, we assign to the temporal graph $\mathbb{G}$ a criticality, by selecting the occurrence with the highest priority in the sequence. 
A criticality node with type in \cref{tab:ontology} (5) is added to the graphs linking to the ego-vehicle node through a $\texttt{criticality}$ relation. We also add a self-loop to the ego-vehicle node representing the scenario's \gls{av} action, with type in \cref{tab:ontology} (4).

\textbf{Seed Graphs. }From the graphs $\mathbb{G}^T$, we construct the temporal representation of the \textit{seed scene graphs}, $\mathbb{\tilde{G}}^T$, where $\mathbb{\tilde{G}}^T = (\mathbb{\tilde{V}}^T, \mathbb{\tilde{E}}^T)$. In particular, we are pruning all edges and relations between ego-vehicle and objects in $\mathbb{G}^T$ to generate $\mathbb{\tilde{G}}^T$. The temporal graphs $\mathbb{\tilde{G}}^T$ contain all the nodes and edges from each \textit{seed scene graph} $\tilde{G_t} \in \mathbb{\tilde{G}}$, where each $\tilde{G_t}$ can be easily retrieved from $\mathbb{\tilde{G}}^T$ using the edge temporal labels $\tau$. During training, these graphs are used as \textit{seeds} for our scene graph generation model which learns to generate the \textit{scenario scene graphs}. To build our agents-structures relations \textit{database}, we also remove the \gls{av} action and criticality components from the temporal \textit{seed scene graphs} before storing them in our database.

\subsection{Temporal Scene Graph Generation}
We design a temporal \gls{gnn} model that learns to generate \textit{scenario scene graphs} by predicting edges connecting the ego-vehicle to agents and static structures, conditioned on a requested \gls{av} action and criticality level. We approach this as a link prediction task, where our model learns to predict edges corresponding to the requested scenario in ontology-augmented scene graphs. Our method builds upon recent link prediction studies, introducing a more refined approach that predicts the existence of multiple temporal edges and relationships between nodes, guided by a structured ontology, instead of relying on exhaustive or broad search strategies that find single links between node pairs in graphs \cite{ wang2019heterogeneous}.

\textbf{Scene Graphs Augmentation.} At the first step of our method, we augment the temporal \textit{seed scene graphs} by adding all possible links and associated edge attributes between the ego-vehicle and the agents and static structures, from our ontology. These edges are added for every timestamp and are labelled with a corresponding temporal label $\tau$ in the augmented $\mathbb{\tilde{G}}^T$.
The links between agents-structures remain unchanged. Our model is explicitly trained to predict only the links involving the ego-vehicle in order to construct the desired \textit{scenario scene graphs} by learning to estimate the likelihood of each edge. The new edges are denoted as $e_e \in \mathbb{\tilde{E}}^T$ with associated feature representations $p_e$. The original agents-structures links are denoted as $e_a \in \mathbb{\tilde{E}}^T$ with associated feature representations $p_a$. We note that both the \textit{criticality} node, connected via the criticality edge, and the \textit{action} edge, represented by a self-loop, are linked to the ego-vehicle. As special cases within the set of edges in $e_a$, both the criticality and action feature vectors are encoded into the ego-vehicle's feature vector, but the corresponding edges are not being predicted. The augmented scene graphs are passed as input to our model.

% To optimise the learning process, we also implement random subsampling during testing to enhance evaluation efficiency while ensuring that any valid edge type from the ontology is retained for accurate inference. We also implement random subsampling during training to optimise the learning process while ensuring that any valid edge type from the ontology is retained for accurate inference.

\textbf{Feature Encoders.} To preserve the structural and spatial information about the nodes, each of them is assigned a sinusoidal positional encoding \cite{attention_all_you_need}, calculated using its unique ID, to capture its global position within the temporal graphs. This allows the network to distinguish between entities of the same class and track their evolving patterns across time.

Each input graph entity is encoded to generate compact feature vector representations. We apply a set of \glspl{mlp} to both the node and edge features of the augmented scene graphs to learn their corresponding embedding representations. In particular, node features are encoded as $z_n = MLP_n(h)$, actors-structures edge features as $z_a = MLP_e(p_a)$, and the ego-vehicle-related edge features as $z_e = MLP_e(p_e)$. We then apply two layers of the \gls{gat} \cite{gat} to aggregate the feature vectors of neighbouring nodes and corresponding edge features in $z_a$ to generate query node embeddings, $z_n' = GAT(z_n, z_a, e_a)$. 

\textbf{Temporal Layers.} The model then sequentially processes the information, guided by the temporal labels $\tau$ in the edges, to capture complex temporal relationships and the evolution of interactions between all scene participants. 
Our temporal encoder receives \textit{node-edge-node} triplets in the input, grouped in sets based on the label $\tau$ in each corresponding edge in the form of $\langle z_{nk}', z_e, z_{nl}' \rangle_\tau$, where $k,l$ are node indices guided by the respective edges. The triplet sets are sequentially passed to the triplet encoder, retaining the order from $\tau$. The encoder consists of a single \gls{mlp} layer that encodes the concatenated feature vectors, $g_{tr} = MLP_{tr}(z_{nk}'|z_e|z_{nl}')_\tau$ for each set sequentially. In between each sequence, we apply a \gls{gcn} layer to further aggregate node feature vectors, capturing both temporal and relational patterns. The updated representations are then propagated to the next step, expressed as $z_n' = GCN(z_n', e_e)$.

\textbf{Link Prediction.} After the last iteration of the temporal module, we apply a Sigmoid function to the triplet embeddings $g_{tr}$ to extract the probability of each triplet, and consequently the likelihood of each edge and edge feature existing in the query \textit{scenario scene graph}, as $\hat{y}=\sigma(g_{tr})$. During training, we use a \gls{bce} loss averaged over all samples as follows:
\begin{align}
        \mathcal{L}_{BCE}(y, \hat{y}) &= - ( y \cdot \log(\hat{y}) + (1-y) \cdot \log(1 - \hat{y}))  \\
        \mathcal{L} &= \frac{1}{N} \sum_{i=1}^{N} \mathcal{L}_{BCE}(y_{i}, \hat{y}_{i})
\end{align}
where $y \in \{0, 1\}$ is the ground-truth label and $\hat{y} \in [0, 1]$ is the predicted triplet likelihood.
In the final step, we apply softmax filtering to retain only the highest probability relation from mutually exclusive edges as defined in the ontology, such as $\{$\texttt{MovingAway}$\}$ and $\{$\texttt{MovingTowards}$\}$. This ensures that the generated graphs remain plausible by preventing contradictions like an entity being classified as \textit{both} moving towards and moving away from the ego-vehicle.

\subsection{Scenario Integration in Simulation}
\textbf{Inference. }
During inference, a user requests a scenario specifying dynamic agents, \gls{av} action, and criticality level. Based on the requested agents, we sample agents-structures relations from our database to build \textit{seed scene graphs}. The specified \gls{av} action and criticality are then reintroduced into the graphs, ensuring they retain a structure similar to the \textit{seed scene graphs} used during training. In particular, we add the AV action as ego-vehicle self-edge and add the criticality as a node linking to the ego-vehicle node. During inference, these graphs are passed in the input of our temporal scene graph generation model which, also constrained by the ontology, generates links corresponding to the requested scenario.

\textbf{Simulation. }The nodes and predicted relationships are then extracted from the graphs at each $\tau$ to build the requested scenarios in CARLA \cite{carla} using the OpenSCENARIO\footnote{\url{asam.net/standards/detail/openscenario/}} standard. The behaviour of the scene participants and the scenario parameters are defined by the scene graphs. We use the relative position and action edges to define the lane and orientation of each agent in the simulation with respect to the ego-vehicle. The initial position of each actor is defined by the scenario's criticality and corresponds to the distance from the ego-vehicle. For each actor, the start and goal positions and orientations are determined by the lane position edges in the graphs, ensuring realistic movement and positioning throughout the scenario.

\section{Experimental Setup}
\subsection{Dataset}
For this work, we used the \gls{road} dataset~\cite{road}, an extension of the Oxford RobotCar Dataset \cite{RobotCarDatasetIJRR}, to train and evaluate our approach, as well as build our agents-structures database.
This dataset is designed to test an \gls{av}’s ability to detect various road events. 
Video sequences were downsampled 5-fold from 12 frames per second (fps) to increase variation across 5-frame sequences, similar to \cite{kochakarn2023explainable}.
% each approximately \SI{2}{\sec} long.
Each sequence corresponds to a specific action of the ego-vehicle, defined as AV Action in \cref{tab:ontology} (4).
We extract the proximity relations between actors and the ego-vehicle by computing and thresholding the box-wise median Relative Inverse Depth (RID) obtained from an off-the-shelf Midas (DPT Large) network~\cite{ranftl2022midas}, with Near Collision ($<5$ $m$), Near ($5-10$ $m$), and Visible ($>10$ $m$).
% We then preprocess the given scene graph labels to generate the temporal scene graphs as described in \cref{sec:methodology:graphextraction}.
We generate $6,771$ temporal graphs grouped by action, where each action group is split into training, validation, and testing sets using a 70-20-10 allocation for balanced evaluation. 

\subsection{Training}

To ensure stable learning and effective generalisation, we carefully tuned hyperparameters and employed several training techniques. For hyperparameters tuning, we used k-fold cross-validation with $k=5$. The dimensions of the MLP layers in our model were configured as follows: $MLP_n:[22, 8, 1]$, $MLP_e:[27, 8, 1]$, $MLP_{tr}:[3, 16, 1]$, and $MLP_{gat}=[64, 128, 256]$. The first \gls{gat} layer used dimensions $[1, 64]$, followed by another \gls{gat} layer $[256, 1]$. Finally, the \gls{gcn} was set with dimensions $[1, 1]$ to capture recurrent node aggregations. This layered structure allowed to process efficiently both node and edge features and generate the required triplet embedding representations. We additionally used a sinusoidal positional encoding of length $2$, the Adam optimiser with $0.01$ learning rate, and weight decay with $\lambda=10^{-5}$. Gradient clipping at $\pm1$ prevented exploding gradients, while Xavier initialisation helped stabilise weight distributions across layers.
%We also employed weight decay with $\lambda=10^{-5}$ to mitigate overfitting. 
% To optimise learning and address any class imbalance, we randomly pruned negative edges during training, resulting in an equal number of positive and negative samples in each class.

\subsection{Baselines}
We evaluated our method against three key baselines.
\begin{enumerate}
    \item F-S LLM: Inspired by ChatScene \cite{zhang2024chatscene}, we evaluated our method against a Few-Shot In-Context Learning (ICL) setup using GPT-4 \cite{openai2024gpt4technicalreport}. We first converted our ontology along with the nodes, edges, and attributes of the \textit{seed scene graphs} in our testing dataset into lists. We then provided the model with these lists and an example response within the prompt, asking it to find plausible relationships between scene entities to construct query scenarios, given our ontology and existing relationships. This comparison assessed the strengths and limitations of LLMs in structured data prediction, particularly in scene graphs.
    \item GRAN$\mathbf{^+}$: We employ the model architecture proposed in GRAN \cite{liao2019gran} and adapt it for our task by incorporating node and edge feature encoders while preserving the attention encoder and GRU. We adjust the prediction mechanism to our triplet prediction task and train the model in our pipeline. Among graph generation models, GRAN is the most comparable to our approach as it predicts links from augmented graphs and its architecture facilitates guiding the generation process through ontology-based constraints.
    \item CC-SGG \cite{ccsgg}: A non-temporal \gls{gnn} model that learns to predict links corresponding to static corner case scene graphs. We adapt it for our task and train the model on our dataset to predict the entire sequence in a single step. This comparison highlights the advantage of our temporal \gls{gnn} method over static \glspl{gnn} in capturing evolving patterns in the input.
\end{enumerate}

%===============================================================================
\section{Results}
\label{sec:result}

\subsection{Classification Performance}
We evaluate our link prediction method using standard classification metrics: F1 score, accuracy, precision, and recall, as seen in \cref{tab:benchmark_diff}. Overall, our model outperforms the baselines achieving the highest \textbf{F1 score} $\mathbf{0.706}$, and \textbf{recall} $\mathbf{0.859}$, and comparable accuracy and precision. Our model achieves the optimal balance in its predictions, effectively minimising false negatives while maintaining strong overall performance.

\begin{table}[h]
\vspace{-5pt}
\centering
\renewcommand{\arraystretch}{1.4}
\begin{tabular}{l|cccc}
                             & \textbf{F1}                      & \textbf{Accuracy}              & \textbf{Precision}             & \textbf{Recall}              \\  \hline
F-S LLM            & 0.218                           & 0.436                                 & 0.212                                  & 0.262               \\  

GRAN$\mathbf{^+}$ & 0.647                           & 0.752                                 & 0.663                                  & 0.704                               \\

CC-SGG   & 0.583    & \textbf{0.801}    & \textbf{0.729}  & 0.510   \\ \hline

Ours & \textbf{0.706}                  & 0.777                        & 0.623                         & \textbf{0.859}                     
\end{tabular}
\caption{Link prediction accuracy across the baselines.}
\vspace{-5pt}

\label{tab:benchmark_diff}
\end{table}

\subsection{Contextual Evaluation}
We additionally conduct a contextual evaluation of the output predictions using the \textit{answer correctness} metric from the \gls{ragas}  framework \cite{es-etal-2024-ragas}, designed to measure the accuracy of an \gls{llm}'s answer to a prompt compared to a ground-truth. To achieve this, we convert the predicted and ground-truth graphs into textual descriptions of the respective scenes and compute the corresponding \textit{answer correctness}, obtaining a score between $0$ and $1$. This metric combines both semantic and factual similarity between the predicted and ground truth descriptions without requiring additional context.
% It combines semantic similarity and factual similarity between the predicted and ground-truth graph descriptions and does not require additional context.
The former captures the overall resemblance between them, while the latter computes their factual overlap, i.e. comparing the individual statements.
To obtain such descriptions of our temporal graphs, we produce a paragraph for each constituent graph, including its timestamp and a statement for each relation. For example, the graph at frame $42$ with relations $\{\langle$\texttt{Pedestrian}, \texttt{Near}, \texttt{EGO}$\rangle,...\}$ becomes \textit{"At time 42: Pedestrian 1 is near the ego-vehicle. ..."}.
This relation-to-statement mapping aligns well with the factual similarity component, providing a contextual evaluation of our predicted graphs' structure.

% Mapping each relation to a statement is particularly in line with the factual similarity component of our metric as it provides a contextual measure of our predicted graphs' structure.

The \textit{answer correctness} results per AV action and criticality are presented in \cref{tab:ragas_actions,tab:ragas_criticalities}. Our model achieves the highest \textit{answer correctness} scores across almost all evaluations, demonstrating its strong ability to generate both accurate and semantically meaningful relations. In contrast, the lower scores of the \gls{llm}-based method justify the benefits of model-based approaches, such as ours, even in the context of semantic understanding.

\begin{table}[h]
\centering
\renewcommand{\arraystretch}{1.4}
\setlength{\tabcolsep}{3pt}
% \begin{tabular}{l|ccccccc|ccc}
\begin{tabular}{l|ccccccc|c}
\multirow{2}{*}{\textbf{Method}} & \multicolumn{7}{c}{\textbf{AV Actions}} \\
% \multirow{2}{*}{\textbf{Method}} & \multicolumn{7}{c|}{\textbf{AV Actions}} \\ & \multicolumn{3}{c}{\textbf{Criticality}} \\ %\cline{2-11} 

  % & Mov & MovL & MovR & OTK & Stop & TurL & TurR & Near & NearCol & Vis \\ \hline
  & Mov & MovL & MovR & Ovtk & Stop & TurnL & TurnR & Avg\\ \hline

% CC-SGG & \textbf{0.783} & 0.784 & 0.902 & \textbf{0.907} & \textbf{0.771} & 0.753 & 0.832 \\% & - & - & - \\ %\hline
F-S LLM & 0.668 & 0.721 & 0.478 & 0.761 & 0.55 & 0.619 & 0.689 & 0.64 \\ % & 0.631 & 0.596 & 0.674 \\
GRAN$\mathbf{^+}$ & 0.735 & 0.823 & 0.667 & 0.715 & 0.763 & 0.733 & 0.878 & 0.746 \\  % & 0.706 & 0.667 & 0.813 \\ %\hline
CC-SGG & 0.709 & 0.615 & 0.814 & 0.732 & \textbf{0.805} & 0.771 & \textbf{0.883} & 0.735 \\ \hline % & - & - & - \\ %\hline \hline

Ours & \textbf{0.758} & \textbf{0.862} & \textbf{0.972} & \textbf{0.776} & 0.769 & \textbf{0.761} & 0.831 & \textbf{0.765}  % & \textbf{0.736} & \textbf{0.68} & \textbf{0.835} \\ %\hline
\end{tabular}
\caption{Contextual evaluation - per \gls{av} action.}
\label{tab:ragas_actions}
\end{table}

This approach, alongside the classification evaluation described above, allowed us to measure not only how well each model performed but also how accurate and contextually appropriate the predictions were.

% \gls{ragas} is usually used to evaluate a model's ability to retrieve relevant and reliable information, process it, and generate a coherent, factual, and accurate answer that addresses the user's query. We employ this measure to evaluate the contextual predictions of our model, similar to the evaluation in \glspl{llm}. High correctness in \gls{ragas} shows that the model can effectively generate a response that is factually accurate and correct.

% We evaluate ...
% Our results demonstrate ....

% \begin{table}[]
% \centering
% \renewcommand{\arraystretch}{1.4}
% \begin{tabular}{l|l}
% \cellcolor[HTML]{FFFFFF} & \textbf{Correctness} \\ \hline
% AV-Move                  &                      \\
% AV-Stop                  &                      \\
% AV-TurnLeft              &                      \\
% AV-TurnRight             &                      \\
% AV-MoveLeft              &                      \\
% AV-MoveRight             &                      \\
% Overtake                 &                      \\ \hline
% All                      &                     
% \end{tabular}
% \caption{Contextual evaluation}
% \label{tab:ragas}
% \end{table}

\begin{table}[h]
\vspace{-10pt}
\centering
\renewcommand{\arraystretch}{1.4}
\setlength{\tabcolsep}{6pt}
\begin{tabular}{l|ccc}
\multirow{2}{*}{\textbf{Method}} & \multicolumn{3}{c}{\textbf{Criticality}} \\

 & \multicolumn{1}{l}{\textbf{Visible}} & \multicolumn{1}{l}{\textbf{Near}} & \multicolumn{1}{l}{\textbf{Near Collision}} \\ \hline

F-S LLM & 0.674 & 0.631 & 0.596 \\
GRAN$\mathbf{^+}$ & 0.735 & \textbf{0.823} & 0.667\\ 
CC-SGG & 0.794 & 0.694 & 0.675 \\ \hline
Ours & \textbf{0.835} & 0.736 & \textbf{0.68}
\end{tabular}
\caption{Contextual evaluation - per scenario criticality.}
\label{tab:ragas_criticalities}
\vspace{-10pt}
\end{table}

\subsection{Scenario Generation}
To evaluate the suitability of our generated simulation scenarios for \gls{av} testing, we assess the performance of $3$ CARLA agents -- \textit{normal, cautious,} and \textit{aggressive} -- provided by the simulator library. Each agent is tested in $30$ randomly selected scenarios with evenly distributed criticality levels.
We compute the Scenario Consistency Rate (SCR) $(\%)$, measuring whether the agents' final states align with each scenario's criticality, as shown in \cref{tab:navigation}.
Our results demonstrate that our method effectively generates diverse scenarios. The behaviour of the CARLA agents in those scenarios mostly leads to situations matching the intended criticality, especially for \textit{visible} and \textit{near} scenarios. Interestingly, the behaviour of the \textit{aggressive} agent consistently led to more dangerous situations than expected, even in lower-criticality scenarios like \textit{visible} and \textit{near}.
We also note a lower SCR percentage for \textit{near collision} scenarios, likely due to the fact that the agents in CARLA have access to the perfect traffic state -- i.e. knowing the state of all vehicles and pedestrians -- making it easier to avoid collisions.

\begin{table}[h]
\vspace{-5pt}
\centering
\renewcommand{\arraystretch}{1.4}
\setlength{\tabcolsep}{6pt}
\begin{tabular}{l|ccc}
\multirow{2}{*}{\textbf{Agents}} & \multicolumn{3}{c}{\textbf{Criticality}} \\
 & \multicolumn{1}{l}{\textbf{Visible}} & \multicolumn{1}{l}{\textbf{Near}} & \multicolumn{1}{l}{\textbf{Near Collision}} \\ \hline
Normal & 70 & 80 & 40 \\
Cautious & 80 & 80 & 50 \\
Aggressive & 60 & 60 & 50
\end{tabular}
\caption{Performance evaluation of $3$ CARLA agents, measured in SCR $(\%)$ for assessing alignment with scenario criticality. Predicted scene graphs define the behaviours and positions of non-ego-vehicle entities, as the ego-vehicle navigates the generated scenario.}
\label{tab:navigation}
\vspace{-15pt}

\end{table}

\begin{figure*}[t]
\vspace{7pt}
\centering
\begin{subfigure}[t]{0.45\textwidth}
    \includegraphics[trim=0 190 0 0,clip,width=\textwidth]{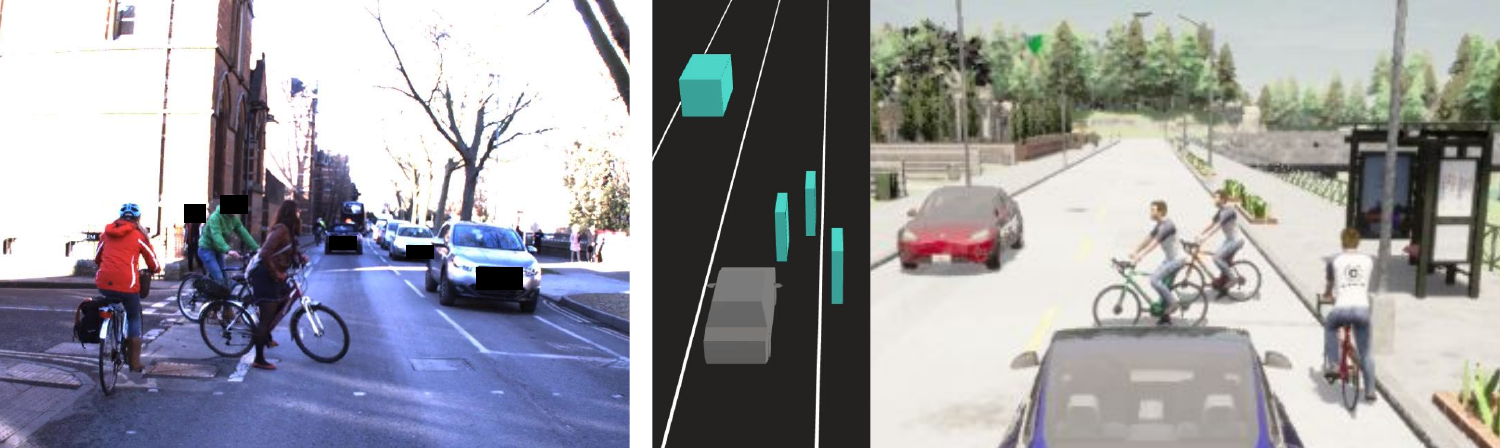}
    \caption{Action: AV-Stop, Criticality: Near Collision}
\end{subfigure}
 \hspace{9pt}
% \hspace{\fill} % maximize horizontal separation
\begin{subfigure}[t]{0.45\textwidth}
    \includegraphics[trim=0 190 0 0,clip,width=\linewidth]{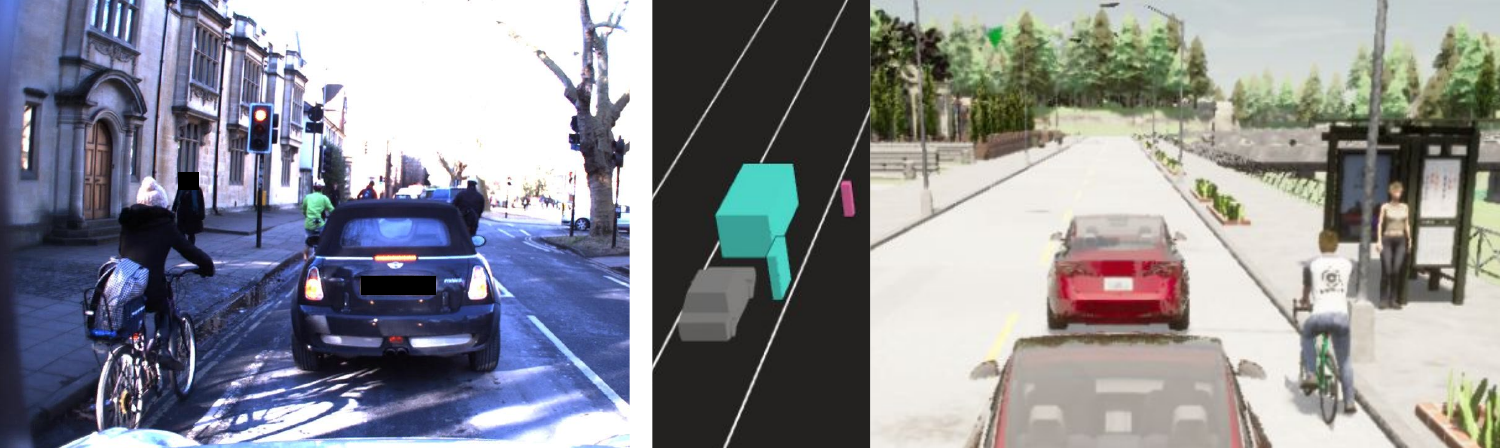}
    \caption{Action: AV-Stop, Criticality: Near}
\end{subfigure}

\bigskip % more vertical separation
\vspace{-5pt}
\begin{subfigure}[t]{0.45\textwidth}
    \includegraphics[trim=0 190 0 0,clip,width=\linewidth]{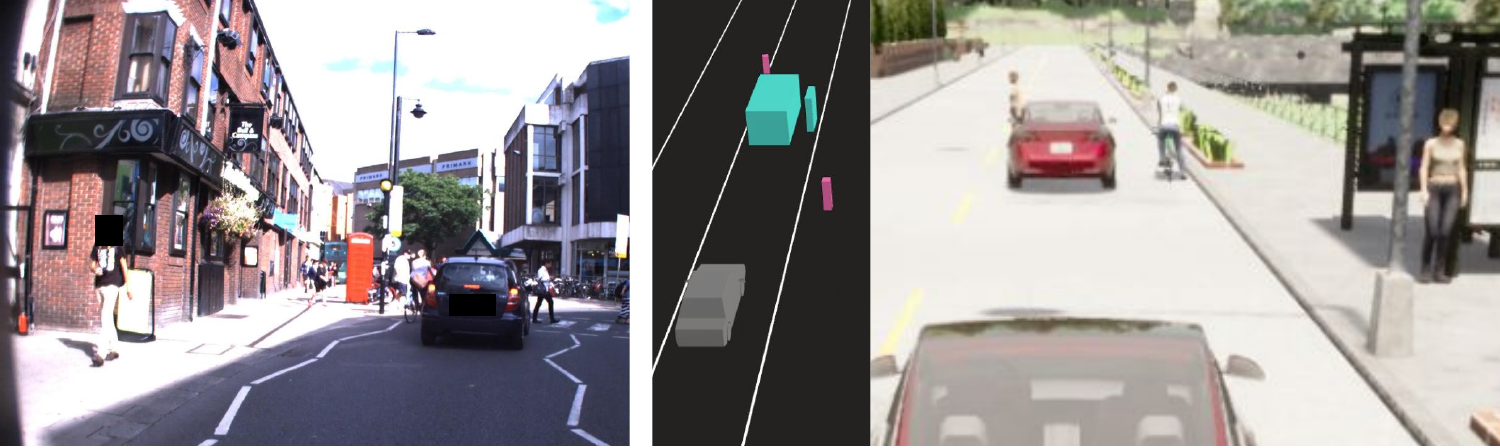}
    \caption{Action: AV-TurnRight, Criticality: Visible }
\end{subfigure}
 \hspace{9pt}
\begin{subfigure}[t]{0.45\textwidth}
    \includegraphics[trim=0 190 0 0,clip,width=\linewidth]{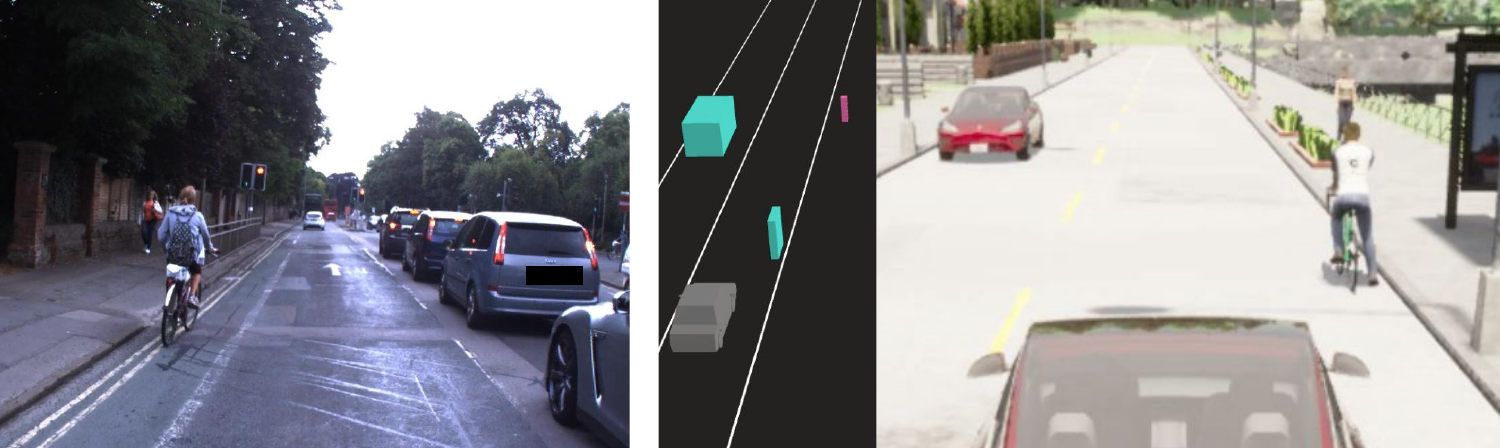}
    \caption{Action: AV-Move, Criticality: Near}
\end{subfigure}
\caption{Qualitative evaluation of the generated scenarios in CARLA using sampled \textit{seed scene graphs} from the \gls{road} dataset~\cite{road} for various actions and criticalities. As the dataset was recorded in a left-hand drive country, we mirrored the topological information to align with the configuration used in CARLA, a right-hand drive environment.}
\label{fig:qualitative}
\vspace{-10pt}
\end{figure*}

\subsection{Qualitative Results}
We qualitatively assess our system's ability to generate novel scenarios from random user inputs. Here, we compare the agent's positions and behaviours in the generated scenarios in CARLA against the agent's positions and behaviours in the scenarios in the \gls{road} dataset from which the sampled \textit{seed scene graphs} originated, depicted in \cref{fig:qualitative} which also illustrates the actors' positions on the map. In our scenarios, actors and associated relational information are successfully generated in simulation and actor behaviour generally aligns with expected criticality levels, as seen in \cref{fig:qualitative} (a,c,d). We observe that, in some cases, the `Near' collision criticality results in more severe situations than expected, e.g. in \cref{fig:qualitative} (b), due to small distance threshold margins for \textit{near} and \textit{near collision} scenarios. To improve distinction, we manually adjust these thresholds in simulation.

%===============================================================================
\section{Ablation}
\label{sec:ablation}
We evaluate several models as candidate triplet encoders, exploring both temporal and non-temporal architectures. 
% Our primary objective is to establish the architecture that maximises the F1 score while achieving a balanced trade-off between precision and recall.

\textbf{Temporal Triplet Encoders. }The goal of this ablation study is to find the optimal model for the temporal triplet encoder. Our results, shown in \cref{tab:model_performance}, indicate that, from the candidate architectures, the GRU with GCN demonstrates promising performance but imbalance between precision and recall. The MLP with GCN encoder consistently outperforms the others and thus we select it for our method.
\begin{table}[h]
    \vspace{-5pt}
    \centering
    \renewcommand{\arraystretch}{1.4}
    \begin{tabular}{l|cccc}
        \textbf{Model} & \textbf{F1} & \textbf{Accuracy} & \textbf{Precision} & \textbf{Recall} \\
        \hline
        GRU      & 0.431 & 0.773 & 0.504 & 0.401 \\
        GRU/GCN  & 0.655 & 0.732 & 0.559 & 0.835 \\
        GAT      & 0.518 & 0.380 & 0.380 & \textbf{1.000} \\ \hline
    Ours (MLP/GCN) & \textbf{0.706} & \textbf{0.777}  & \textbf{0.623}  & 0.859
    \end{tabular}
    \caption{Ablation study for various triplet encoders.}
    \label{tab:model_performance}
    \vspace{-10pt}
\end{table}

\textbf{Non-Temporal Triplet Encoders. }We also assess the performance of non-temporal triplet encoders for this task. Our temporal encoder achieves higher F1 score than non-temporal encoders, indicating a well-balanced trade-off between precision and recall as seen in \cref{tab:ablation_nonseq}. We also note that the non-temporal MLP, which also showed promising results, is very similar to the encoder used in CC-SGG, and, as such, its performance was previously evaluated in \cref{sec:result}.

\begin{table}[h]
\vspace{-12pt}
\centering
\renewcommand{\arraystretch}{1.4}
\begin{tabular}{l|cccc}
    & \multicolumn{1}{l}{\textbf{F1}} & \multicolumn{1}{l}{\textbf{Accuracy}} & \multicolumn{1}{l}{\textbf{Precision}} & \multicolumn{1}{l}{\textbf{Recall}} \\ \hline
GRU & 0.319 & 0.420 & 0.246 & 0.502 \\
GAT & 0.518 & 0.380 & 0.380 & \textbf{1.000} \\
MLP & 0.675 & \textbf{0.784} & \textbf{0.644} & 0.745  \\ \hline
Ours & \textbf{0.706}                  & 0.777      & 0.623  & 0.859
\end{tabular}
\caption{Non-Temporal Triplet Encoders compared to our temporal encoder.}
\label{tab:ablation_nonseq}
\vspace{-10pt}
\end{table}

% \begin{figure}[h]
% \vspace{-10pt}
% \centering
% \includegraphics[width=0.75\columnwidth]{images/ablation_icra.png}
%  \caption{Ablation study for various triplet encoders.}
% \label{fig:ablation}
% \vspace{-15pt}
% \end{figure}

\textbf{Positional Encodings. }We also evaluate the impact of the positional encodings in our graphs. For this, we employ a dataset containing actions with lower variability, \textit{AV-Move, AV-Stop, AV-TurnLeft, AV-TurnRight}, and build graphs with a simpler architecture, where same-type actors were mapped to the same node in the graph, and without positional encodings. We evaluate the performance of all baselines on these graphs in \cref{tab:benchmark_easy}.
While our model achieves strong recall, effectively identifying true positive instances, its low F1 score indicates unbalanced predictions. Across almost all metrics, baseline models exhibit consistently poor performance.
These results support the argument for more descriptive node features to improve differentiation and tracking across actors. 
\begin{table}[h]
\centering
\vspace{-10pt}
\renewcommand{\arraystretch}{1.4}
\begin{tabular}{l|cccc}
                              & \textbf{F1}                      & \textbf{Accuracy}              & \textbf{Precision}             & \textbf{Recall}              \\ \hline

F-S LLM & 0.207 & 0.427 & 0.204 & 0.249 \\ 
GRAN$\mathbf{^+}$ & 0.524 & 0.605 & 0.506 & 0.605 \\ 
CC-SGG & 0.610 & \textbf{0.795} & \textbf{0.679} & 0.586 \\ \hline
Ours & \textbf{0.669} & 0.702 & 0.587 & \textbf{0.818} 
\end{tabular}
\caption{Performance without positional encodings.}
\label{tab:benchmark_easy}
\vspace{-15pt}
\end{table}
%===============================================================================

\section{Conclusion}
\label{sec:conclusion}
Our proposed methodology effectively generates diverse scenarios in simulation by modelling complex semantic interactions through temporal scene graphs. We evaluated the accuracy and contextual relevance of our model's predictions, achieving superior performance to the baseline methods. Additionally, we assessed the performance of $3$ navigation agents in our generated scenarios, demonstrating their suitability as testing environments. Our results further indicate that our approach can generate scenarios \textit{on-demand}, achieving the desired criticality levels. Overall, our method establishes a robust framework for dynamic scenario generation, leveraging temporal graphs and \glspl{gnn} to enhance \gls{av} testing and evaluation in diverse scenarios.
% ----------------------------
\section*{Acknowledgements}
We thank Valentina Mu\c{s}at for reviewing the paper prior to submission, and Harry Way and Pawit Kochakarn for their work on the early stages of the scene graph pipelines.

% ----------------------------
\bibliographystyle{IEEEtran}
\bibliography{biblio}

\end{document}